\pdfoutput=1
\documentclass[11pt]{article}
\usepackage[a4paper,margin=2.4cm]{geometry}
\usepackage[utf8]{inputenc}
\usepackage[T1]{fontenc}
\usepackage{lmodern}
\usepackage{graphicx,booktabs,longtable,amsmath,amssymb,url}
\usepackage[numbers,sort&compress]{natbib}
\usepackage{placeins}
\usepackage{titlesec}
\usepackage{caption}
\usepackage[hidelinks,
            pdftitle={Target leakage, not model class, explains reported accuracy in survey-based cardiovascular screening},
            pdfauthor={Raad Bin Tareaf; Murad Al-Rajab; Samia Loucif; Samer Ellaham; Cedric Schmitz},
            pdfsubject={Machine learning evaluation for cardiovascular screening on national health survey data},
            pdfkeywords={target leakage; clinical prediction; glass-box models; tabular foundation models; algorithmic fairness; conformal prediction; BRFSS}
           ]{hyperref}

\titleformat{\section}{\normalfont\large\bfseries}{\thesection.}{0.6em}{}
\titleformat{\subsection}{\normalfont\bfseries}{}{0pt}{}
\titleformat{\subsubsection}{\normalfont\itshape}{}{0pt}{}
\setcitestyle{square,comma}
\newcommand{\suppfig}[1]{}
\begin{document}

\begin{center}
{\LARGE\bfseries Target leakage, not model class, explains reported\\[3pt]
accuracy in survey-based cardiovascular screening}\\[8pt]
{\large a leakage-tiered audit of glass-box and tabular foundation models}\\[16pt]

{\large Raad Bin Tareaf\,$^{1,2,*}$\quad Murad Al-Rajab\,$^{3}$\quad
Samia Loucif\,$^{4}$\\[4pt]
Samer Ellaham\,$^{5}$\quad Cedric Schmitz\,$^{1}$}\\[12pt]

{\small
$^{1}$Data Science and AI Cluster, XU Exponential University of Applied Sciences, Potsdam, Germany\\
$^{2}$German University of Digital Science, Potsdam, Germany\\
$^{3}$College of Engineering, Abu Dhabi University, Abu Dhabi, United Arab Emirates\\
$^{4}$College of Technological Innovation, Zayed University, Abu Dhabi, United Arab Emirates\\
$^{5}$Cleveland Clinic Hospital, Abu Dhabi, United Arab Emirates\\[6pt]
$^{*}$Correspondence: \href{mailto:r.bintareaf@xu-university.de}{r.bintareaf@xu-university.de}\quad
ORCID: \href{https://orcid.org/0000-0002-2804-3243}{0000-0002-2804-3243}}
\end{center}
\vspace{8pt}

\section*{ABSTRACT}

\textbf{Objective.} Cardiovascular screening models trained on national
health surveys routinely report areas under the receiver operating
characteristic curve (AUROC) near 0.89. We asked whether that accuracy
reflects learning or target leakage, whether tabular foundation models
change the answer, and whether the properties deployment requires survive
joint examination.

\textbf{Materials and Methods.} We benchmarked ten classifiers spanning
linear, tree-ensemble, neural, glass-box and tabular foundation classes for
prevalent myocardial infarction in 442\,067 respondents of the 2022
Behavioral Risk Factor Surveillance System, across five feature tiers of
decreasing leakage risk. Each was audited for discrimination,
calibration, fairness at an explicit screening threshold, conformal
coverage, explanation faithfulness and inference cost, then applied ---
models and thresholds frozen --- to 430\,755 respondents of 2023.

\textbf{Results.} Removing two post-diagnostic features cost every model
0.049--0.051 AUROC, collapsing the field into a 0.0045-wide band. The
glass-box explainable boosting machine was non-inferior to every
alternative within a pre-specified 0.005 margin while scoring the cohort
roughly $10^4$ times faster than the strongest foundation model. One
threshold detected 75.4\% of women's infarctions against 89.0\% of men's;
editing the model's shape functions reduced the gap to 0.010. Marginal
conformal prediction gave 0.86 coverage to men and 0.82 to adults over 60;
Mondrian calibration repaired every stratum. Frozen models transported
within 0.002 AUROC.

\textbf{Discussion.} Reported headroom in this literature is a property of
the feature set, not the learner. Transparency cost nothing measurable and
made fairness repair and uncertainty conditioning directly auditable.

\textbf{Conclusion.} Evaluation practice, not model capacity, is the
binding constraint.

\medskip
\noindent\textbf{Keywords.} target leakage; clinical prediction models;
glass-box models; tabular foundation models; algorithmic fairness; conformal
prediction; Behavioral Risk Factor Surveillance System

\medskip\hrule\medskip

\section*{BACKGROUND AND SIGNIFICANCE}

Cardiovascular disease remains the leading cause of death worldwide, and
population health surveys are an attractive substrate for screening
triage: they are cheap, national in scope, and require no clinical
encounter. A large applied literature has grown on one such substrate, the
Centers for Disease Control and Prevention's Behavioral Risk Factor
Surveillance System (BRFSS), and on its widely redistributed derivatives.
Reported discrimination in that literature clusters remarkably tightly
around an AUROC of 0.89, a figure now repeated across dozens of studies and
several public-facing tools.

That figure deserves scrutiny. Target leakage --- the presence of
predictors that are consequences of the outcome rather than antecedents of
it --- inflates apparent performance without conferring any screening
ability, and is the single most common defect in machine-learning-based
science.\cite{davis2023leakage} Diagnosis codes recorded in the same
encounter as the outcome have been shown to produce exactly this
failure,\cite{ramadan2025leakage} and single-pipeline corrections on
cardiovascular survey data point the same way.\cite{eltawil2026smote} What
the literature lacks is a dose--response measurement: how much of the
reported accuracy survives when leakage-prone predictors are removed in
controlled steps, and whether the answer depends on the learner.

The question has acquired new urgency. Tabular foundation models perform
competitive classification with no task-specific
training,\cite{hollmann2025tabpfn} and in-context learning now extends to
training sets of hundreds of thousands of rows.\cite{qu2025tabicl} Both are
natural candidates for national-survey screening, yet published clinical
evaluations emphasise discrimination on modest cohorts and leave open how
they behave under leakage stress, subgroup audit, conformal calibration and
temporal shift --- and what their inference cost implies at population
scale. At the opposite pole, glass-box generalized additive models have
matched black-box accuracy on clinical tabular data for a
decade,\cite{caruana2015intelligible,lou2013ga2m} and their editable
additive structure has been proposed as a vehicle for interactive model
repair.\cite{wang2022gamchanger} Editing has never been quantified as a
fairness intervention against standard mitigation baselines, nor have
glass-box explanations been used as ground truth to measure how much
post-hoc explainers actually recover.\cite{lundberg2017shap,ribeiro2016lime}

These gaps compound. Fairness in this literature is usually reported
threshold-free, so the operating point at which harm would occur is never
named. Uncertainty is rarely quantified at all, and where conformal
prediction is applied its marginal guarantee is seldom conditioned on the
groups that matter.\cite{gibbs2023conditional,sun2025fairicp} Temporal
validation, when present, permits refitting, which silently tests a
different claim from the one deployment makes. No study subjects one cohort
to all of these audits at once, so the field cannot say whether its
celebrated numbers, its fairness anecdotes and its uncertainty claims
describe the same models under the same conditions.
Table~\ref{tab:positioning} situates this work against the closest prior
studies.

\begin{table}[!htb]\centering
\caption{Positioning against the closest prior work. ``Tabular FM at
scale'' denotes in-context evaluation with a $\geq\!10^5$-row training
context.}
\label{tab:positioning}
{\scriptsize\setlength{\tabcolsep}{3pt}
\begin{tabular}{@{}lcccccccc@{}}
\toprule
 & \shortstack{Leakage\\tiers} & \shortstack{Tabular FM\\at scale}
 & \shortstack{Fairness @\\explicit thr.} & \shortstack{Glass-box\\repair}
 & \shortstack{Group-cond.\\conformal} & \shortstack{Faithfulness\\vs exact}
 & \shortstack{Temporal\\ext.\ valid.} & \shortstack{Open\\pipeline}\\
\midrule
TabPFN v2 (2025)~\cite{hollmann2025tabpfn} & -- & (\checkmark) & -- & -- & -- & -- & -- & \checkmark\\
TabICL (2025)~\cite{qu2025tabicl} & -- & \checkmark & -- & -- & -- & -- & -- & \checkmark\\
Eltawil \emph{et al.} (2026)~\cite{eltawil2026smote} & (\checkmark) & -- & -- & -- & -- & -- & -- & --\\
Davis \emph{et al.} (2023)~\cite{davis2023leakage} & (\checkmark) & -- & -- & -- & -- & -- & -- & --\\
Ramadan \emph{et al.} (2025)~\cite{ramadan2025leakage} & (\checkmark) & -- & -- & -- & -- & -- & -- & --\\
Caruana \emph{et al.} (2015)~\cite{caruana2015intelligible} & -- & -- & -- & \checkmark & -- & -- & -- & --\\
GAM Changer (2022)~\cite{wang2022gamchanger} & -- & -- & -- & (\checkmark) & -- & -- & -- & \checkmark\\
FAIM (2024)~\cite{faim2024} & -- & -- & \checkmark & -- & -- & -- & -- & --\\
Gibbs \emph{et al.} (2025)~\cite{gibbs2023conditional} & -- & -- & -- & -- & \checkmark & -- & -- & --\\
FairICP (2025)~\cite{sun2025fairicp} & -- & -- & (\checkmark) & -- & \checkmark & -- & -- & --\\
Kwon \emph{et al.} (2026)~\cite{kwon2026conformal} & -- & -- & -- & -- & (\checkmark) & -- & (\checkmark) & --\\
\midrule
\textbf{This work} & \checkmark & \checkmark & \checkmark & \checkmark & \checkmark & \checkmark & \checkmark & \checkmark\\
\bottomrule
\end{tabular}

\smallskip
{\scriptsize \checkmark\ addressed as a study objective; (\checkmark)
partial or methodological precursor; -- not an objective of the cited
work. ``Tabular FM at scale'': in-context evaluation with
$\geq\!10^5$-row training context.}
}
\end{table}
\FloatBarrier

\section*{OBJECTIVE}

We set out to (1) measure how much of the reported discrimination in
survey-based cardiovascular screening is attributable to target leakage
rather than to model capacity, across model classes; (2) test whether a
transparent glass-box model is statistically non-inferior to tuned
gradient boosting and to large-context tabular foundation models once
leakage-prone predictors are removed, and at what inference cost; and
(3) determine whether calibration, subgroup equity at an explicit
screening threshold, group-conditional uncertainty and year-to-year
transportability hold simultaneously for the resulting models.

\section*{MATERIALS AND METHODS}

\subsection*{Data and cohorts}
We used the 2022 BRFSS Landline and Cellular Telephone file for
development and the 2023 file for temporal external
validation.\cite{cdc2022brfss,cdc2023brfss} The outcome was self-reported
ever-diagnosed myocardial infarction (CVDINFR4), a prevalent, surviving,
diagnosis-aware phenotype. Respondents with a missing outcome were
excluded, leaving 442\,067 of 445\,132 raw 2022 records (25\,108 cases,
5.68\%) and 430\,755 for 2023 (5.44\%). Item missingness was retained
natively rather than imputed or dropped: a complete-case design would have
discarded 42.9\% of 2022 and 98.7\% of 2023 respondents, the latter because
two questionnaire items moved to state-optional modules. Cohort
construction, variable mapping and cohort composition are given in
Supplementary Figures~1--2 and Supplementary Table~1.

\subsection*{Leakage-tiered feature design}
\begin{figure}[!htb]\centering
\includegraphics[width=\linewidth]{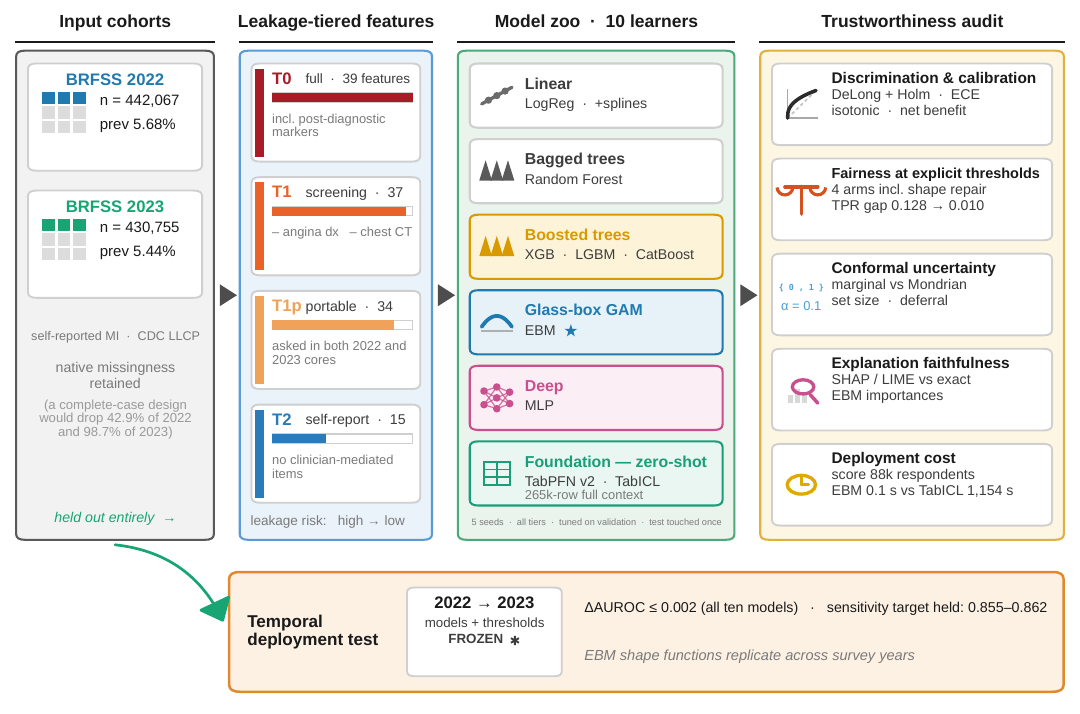}
\caption{\textbf{Study framework.} Two full BRFSS cycles feed a leakage-tiered feature design (T0 $\supset$ T1 $\supset$ T1-portable $\supset$ T2; the finer T1-ns ablation additionally removes prior stroke from T1 and is reported for all ten models in Table~\ref{tab:benchmark}). Ten learners are trained under one pre-specified protocol and subjected to a five-part trustworthiness audit; all 2022 models and their 2022 operating thresholds are then applied frozen to the entire 2023 cycle.}
\label{fig:framework}
\end{figure}

Thirty-nine analytic predictors were assigned to five tiers of decreasing
leakage risk, nested as T0 $\supset$ T1 $\supset$ T1-portable $\supset$ T2,
with T1-ns a sibling ablation of T1 (Figure~\ref{fig:framework}; Supplementary
Table~2).

\FloatBarrier
 \textbf{T0} (39 features) is the full set used by prior work on
this data family and retains two direct post-diagnostic markers: a prior
angina or coronary heart disease diagnosis, and a chest computed
tomography scan. \textbf{T1} (37) removes both and is the primary
screening tier. \textbf{T1-ns} removes prior stroke as well, isolating the
strongest remaining comorbidity proxy. \textbf{T1-portable} (34) restricts
T1 to items asked in both the 2022 and 2023 core questionnaires, and is
the tier used for temporal validation. \textbf{T2} (15) retains only
self-reportable items with no clinician-mediated content. Tier membership
was fixed before any model was fitted.

\subsection*{Models and protocol}
Ten classifiers spanned linear models, bagged and boosted tree ensembles,
a glass-box additive model, a neural network and tabular foundation models:
$\ell_2$-regularized logistic regression and a natural-spline variant
(linear); random forest (bagged trees); XGBoost, LightGBM and CatBoost
(boosted trees);\cite{breiman2001rf,chen2016xgboost,ke2017lightgbm,prokhorenkova2018catboost}
the explainable boosting machine (glass-box additive, the reference model
for all comparisons);\cite{nori2019interpretml} a multilayer perceptron
(neural); and two tabular foundation models applied zero-shot --- TabPFN~v2
with eight 10\,000-row subsampled contexts and TabICL with the full
265\,240-row training context.\cite{hollmann2025tabpfn,qu2025tabicl}

Each tier was run with five seeds under stratified 60/20/20
train/validation/test splits. Hyperparameters were tuned on validation
only, using Optuna;\cite{akiba2019optuna} the test partition was read once
per cell and no test quantity informed any modelling choice. Classical
models used class-weighted training; foundation models were used as
published, without weighting or inference optimisation. Full search spaces,
software versions and hardware are given in Supplementary Methods.

\subsection*{Evaluation}
Discrimination was AUROC and area under the precision--recall curve.
Non-inferiority of the glass-box model was pre-specified at a margin
$\delta=0.005$ AUROC and tested with paired DeLong statistics under
Holm correction, with two-sided equivalence assessed by the
two-one-sided-tests procedure.\cite{delong1988,holm1979} Calibration was
expected calibration error, Brier score and calibration slope, before and
after isotonic recalibration fitted on validation
predictions;\cite{zadrozny2002isotonic} clinical utility was assessed by
decision-curve net benefit across the 1--30\% threshold
range.\cite{vickers2006dca}

A screening threshold was fixed on validation at sensitivity $\geq0.85$ and
all fairness quantities reported at that operating point, with the
true-positive-rate (TPR) gap across sex as the primary disparity metric.
Four mitigation families shared one selection objective, equalizing
validation TPR at the level attained overall at the global threshold:
per-group thresholds; Kamiran--Calders preprocessing
reweighing;\cite{kamiran2012reweighing} glass-box \emph{shape repair}, in
which the sex main effect and all sex-involving interaction terms are
zeroed or attenuated; and shape repair with per-group intercept
equalization. Every edit is a reversible function of the fitted model and
is exported as a machine-readable edit log.

Uncertainty used split conformal prediction at
$\alpha=0.1$,\cite{vovk2005conformal} comparing marginal calibration with
Mondrian calibration within sex and within sex$\times$age strata, and
reporting coverage, deferral (ambiguous $\{0,1\}$ sets) and empty-set
rates. Explanation faithfulness scored TreeSHAP, KernelSHAP and LIME
against the glass-box model's exact global importances by Kendall $\tau$
and top-10 Jaccard overlap across explanation budgets. Inference cost was
wall-clock time on an identical 88\,413-respondent scoring workload.

For temporal validation, all 2022 T1-portable models \emph{and} their 2022
validation-selected thresholds were applied unchanged to the entire 2023
cycle, with no refitting and no re-thresholding. Because BRFSS oversamples
by design, every frozen model was additionally re-evaluated under the
survey design weights. Reporting follows TRIPOD+AI.\cite{collins2024tripod}

\section*{RESULTS}

\subsection*{Two post-diagnostic features account for the reported headroom}
\begin{figure}[!htb]\centering
\includegraphics[width=\linewidth]{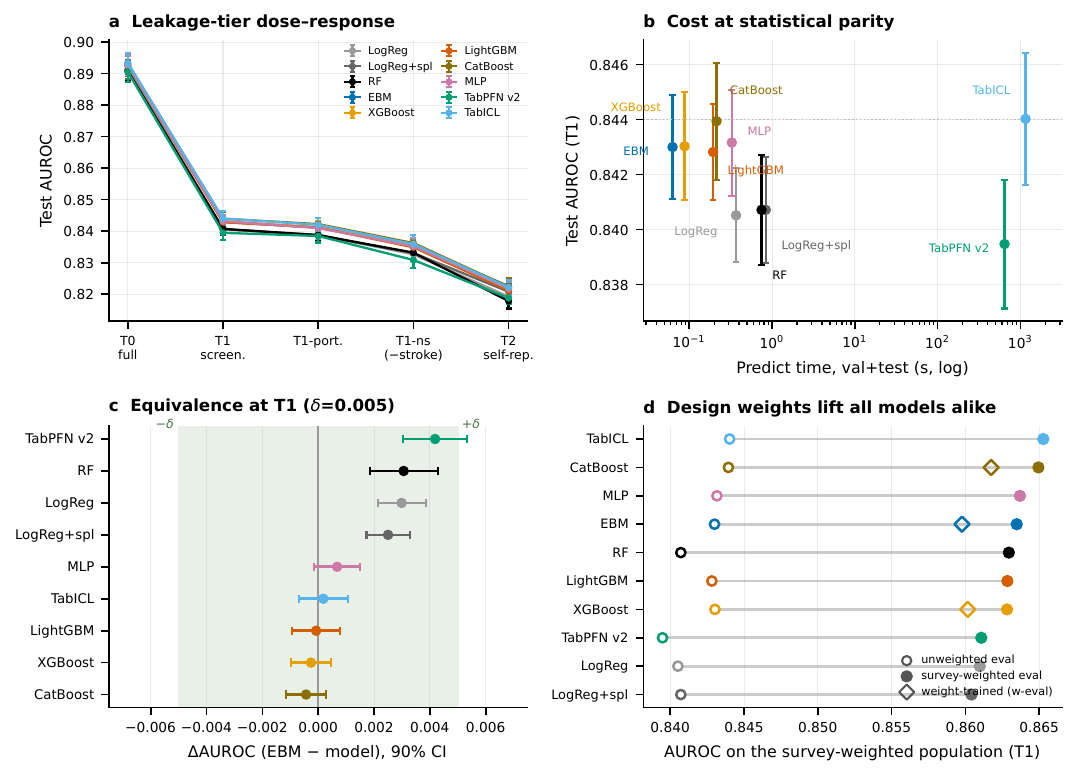}
\caption{\textbf{Leakage, cost, equivalence and design weights.} (\textbf{a}) Test AUROC (mean $\pm$ SD, five seeds) across leakage tiers, ordered by impact. (\textbf{b}) Accuracy versus prediction time on the validation and test workload, logarithmic scale. (\textbf{c}) Non-inferiority forest at T1: difference in AUROC (glass-box minus comparator) with 90\% confidence intervals against the pre-specified $\pm0.005$ band. (\textbf{d}) Survey design weights lift all models nearly uniformly; open diamonds denote weight-informed training.}
\label{fig:leakage}
\end{figure}

On the full T0 feature set, every one of the ten models reached a test
AUROC between 0.8905 and 0.8933 (Table~\ref{tab:benchmark};
Figure~\ref{fig:leakage}a), reproducing the $\sim$0.89 regime of the prior
literature regardless of model class. Removing the two direct
post-diagnostic markers (tier T1) cost 0.049--0.051 AUROC in every model,
near-uniformly, and collapsed the entire zoo into a band 0.0045 wide
(0.8395--0.8440). Removing prior stroke as well (T1-ns) cost a further
0.0075--0.0086, with re-convergence into a 0.0053-wide band. Restricting to
self-report-only predictors (T2) cost a further $\sim$0.021. The uniformity
of the decline across every model class locates the reported headroom in
the feature set, not the learner.

\FloatBarrier

\begin{table}[!htb]\centering
\caption{Main benchmark: test AUROC and area under the precision--recall
curve by leakage tier (mean $\pm$ SD over five seeds).}
\label{tab:benchmark}
{\footnotesize\setlength{\tabcolsep}{3pt}
\begin{tabular}{@{}lccccc@{}}
\toprule
 & \multicolumn{5}{c}{\textbf{AUROC} (mean $\pm$ SD, 5 seeds)}\\
\cmidrule(l){2-6}
Model & T0 (full) & T1 (screen.) & T1-ns & T1-port. & T2 (self-rep.)\\
\midrule
LogReg & 0.8911$\pm$0.0029 & 0.8405$\pm$0.0017 & 0.8326$\pm$0.0021 & 0.8389$\pm$0.0017 & 0.8192$\pm$0.0019 \\
LogReg + splines & 0.8912$\pm$0.0030 & 0.8407$\pm$0.0019 & 0.8331$\pm$0.0023 & 0.8386$\pm$0.0017 & 0.8208$\pm$0.0025 \\
Random Forest & 0.8907$\pm$0.0031 & 0.8407$\pm$0.0020 & 0.8332$\pm$0.0021 & 0.8388$\pm$0.0018 & 0.8178$\pm$0.0024 \\
EBM & 0.8927$\pm$0.0032 & 0.8430$\pm$0.0019 & 0.8354$\pm$0.0023 & 0.8411$\pm$0.0017 & 0.8225$\pm$0.0027 \\
XGBoost & 0.8930$\pm$0.0032 & 0.8430$\pm$0.0020 & 0.8353$\pm$0.0023 & 0.8413$\pm$0.0019 & 0.8223$\pm$0.0024 \\
LightGBM & 0.8925$\pm$0.0032 & 0.8428$\pm$0.0017 & 0.8349$\pm$0.0022 & 0.8412$\pm$0.0016 & 0.8211$\pm$0.0022 \\
CatBoost & 0.8933$\pm$0.0032 & 0.8439$\pm$0.0021 & 0.8362$\pm$0.0026 & 0.8423$\pm$0.0019 & 0.8224$\pm$0.0026 \\
MLP & 0.8927$\pm$0.0030 & 0.8432$\pm$0.0019 & 0.8352$\pm$0.0025 & 0.8410$\pm$0.0019 & 0.8218$\pm$0.0021 \\
TabPFN v2$^{\dagger}$ & 0.8905$\pm$0.0032 & 0.8395$\pm$0.0023 & 0.8309$\pm$0.0027 & 0.8385$\pm$0.0021 & 0.8188$\pm$0.0018 \\
TabICL & 0.8933$\pm$0.0032 & 0.8440$\pm$0.0024 & 0.8359$\pm$0.0029 & 0.8420$\pm$0.0023 & 0.8222$\pm$0.0024 \\
\midrule
 & \multicolumn{5}{c}{\textbf{AUPRC} (mean $\pm$ SD; prevalence 0.0568)}\\
\cmidrule(l){2-6}
LogReg & 0.415$\pm$0.003 & 0.260$\pm$0.002 & 0.237$\pm$0.002 & 0.257$\pm$0.001 & 0.213$\pm$0.000 \\
LogReg + splines & 0.417$\pm$0.004 & 0.261$\pm$0.002 & 0.238$\pm$0.003 & 0.256$\pm$0.002 & 0.215$\pm$0.002 \\
Random Forest & 0.412$\pm$0.005 & 0.255$\pm$0.002 & 0.233$\pm$0.002 & 0.251$\pm$0.002 & 0.210$\pm$0.001 \\
EBM & 0.421$\pm$0.006 & 0.263$\pm$0.002 & 0.239$\pm$0.003 & 0.259$\pm$0.001 & 0.218$\pm$0.002 \\
XGBoost & 0.428$\pm$0.006 & 0.264$\pm$0.002 & 0.239$\pm$0.002 & 0.260$\pm$0.002 & 0.218$\pm$0.001 \\
LightGBM & 0.424$\pm$0.005 & 0.263$\pm$0.002 & 0.239$\pm$0.002 & 0.260$\pm$0.002 & 0.216$\pm$0.001 \\
CatBoost & 0.427$\pm$0.006 & 0.266$\pm$0.002 & 0.243$\pm$0.002 & 0.263$\pm$0.001 & 0.219$\pm$0.002 \\
MLP & 0.426$\pm$0.004 & 0.264$\pm$0.002 & 0.241$\pm$0.002 & 0.260$\pm$0.002 & 0.218$\pm$0.001 \\
TabPFN v2$^{\dagger}$ & 0.422$\pm$0.005 & 0.259$\pm$0.002 & 0.235$\pm$0.002 & 0.256$\pm$0.001 & 0.213$\pm$0.000 \\
TabICL & 0.429$\pm$0.006 & 0.265$\pm$0.002 & 0.241$\pm$0.003 & 0.262$\pm$0.002 & 0.217$\pm$0.001 \\
\bottomrule
\end{tabular}

\smallskip
{\scriptsize $^{\dagger}$TabPFN v2 evaluated under its 8$\times$10k
subsampled-context protocol throughout. T1-ns removes prior stroke from
T1 via the release's pre-specified switch.}
}
\end{table}
\FloatBarrier

\subsection*{The glass-box model is statistically equivalent to every alternative}
At T1 the explainable boosting machine reached 0.8430\,$\pm$\,0.0019
against CatBoost 0.8439, XGBoost 0.8430, the multilayer perceptron 0.8432
and full-context TabICL 0.8440. Under the pre-specified margin
$\delta=0.005$ it was non-inferior to every comparator at every tier
(Holm-adjusted $p<0.001$ throughout), and two-sided equivalence held
against all nine comparators at T0 and T1-portable and eight of nine at T1
(Figure~\ref{fig:leakage}c). The sole exception favoured the glass-box
model: subsampled-context TabPFN~v2 trailed it by 0.0042 (90\% CI
0.0031--0.0053), a gap attributable to the context protocol rather than the
model family. Tree-based comparators correlated with the glass-box
predictions at 0.986--0.991 and the foundation models at 0.83, widening the
latter's paired standard errors as expected.

Parity in accuracy concealed an extreme asymmetry in compute. At T1 the
glass-box model needed 0.1\,s to score the 88\,413-respondent test
partition; TabICL required 3.5\,s to ingest its context but 1153.8\,s of
GPU inference for the identical workload, and TabPFN~v2 642.5\,s
(Figure~\ref{fig:leakage}b). Scaled to a national cohort, this separates
sub-second scoring from tens of GPU-minutes per pass.

\subsection*{Discrimination parity, calibration divergence}
\begin{figure}[!htb]\centering
\includegraphics[width=.95\linewidth]{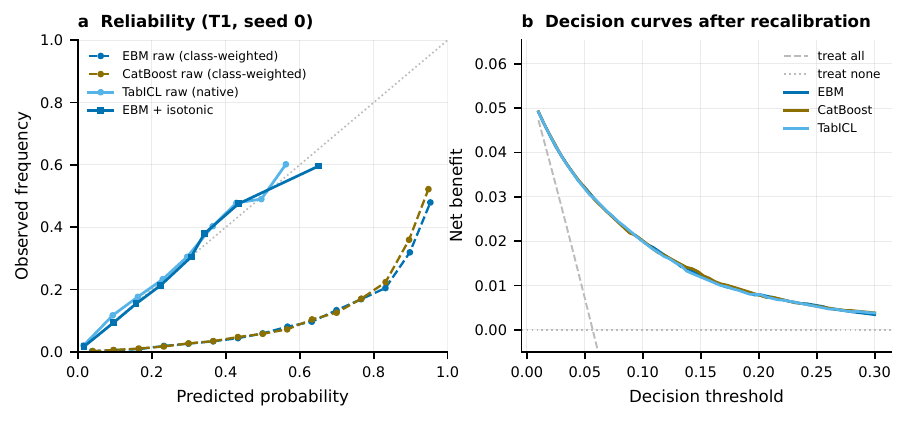}
\caption{\textbf{Calibration and clinical utility at tier T1.} (\textbf{a}) Reliability curves before and after isotonic recalibration. (\textbf{b}) Decision-curve net benefit against treat-all and treat-none strategies across the 1--30\% threshold range.}
\label{fig:calib}
\end{figure}

Raw probability quality separated the classes where discrimination could
not (Figure~\ref{fig:calib}a). Class-weighted classical models were
severely miscalibrated in absolute risk (expected calibration error
0.234--0.291) while their calibration slopes remained near ideal
(0.91--1.20) --- an arithmetic consequence of balanced weighting, not a
property of model class, as confirmed by refitting without class weights
(error 0.0031 and 0.0019 for the glass-box model and CatBoost). The two
foundation models, which apply no class weighting, were natively calibrated
(error 0.010). Isotonic recalibration restored every model to error
$\leq0.005$ without altering discrimination, after which all competitive
models showed positive net benefit across the 1--30\% threshold range
(Figure~\ref{fig:calib}b).

\FloatBarrier

\subsection*{A single threshold under-detects women; glass-box repair closes the gap}
\begin{figure}[!htb]\centering
\includegraphics[width=.62\linewidth]{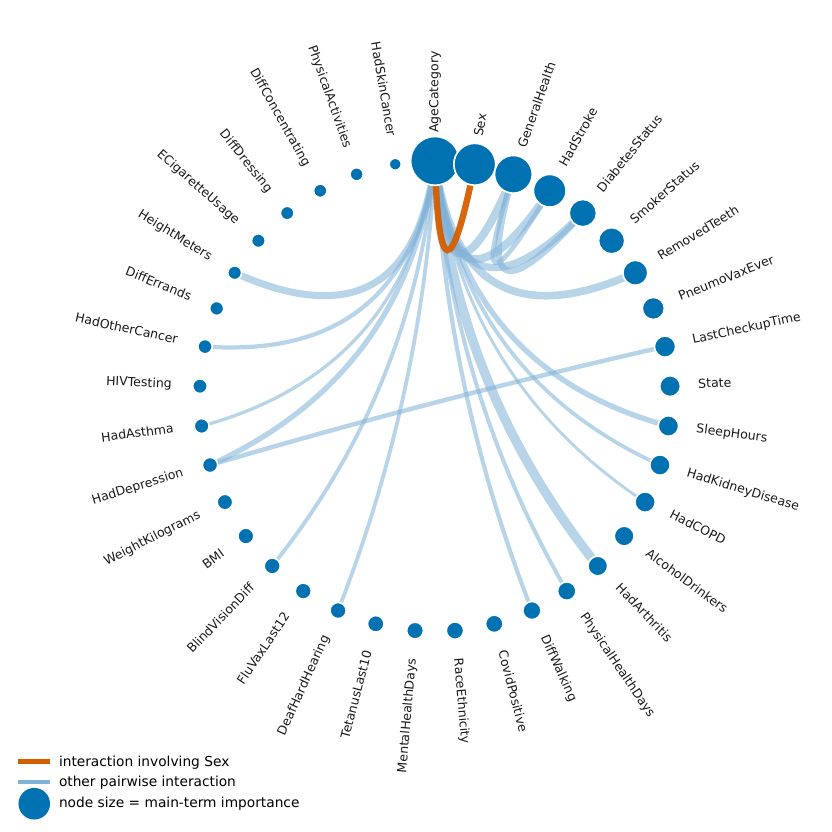}
\caption{\textbf{Learned pairwise-interaction structure of the glass-box model.} Tier T1, seed 0. Nodes are main-effect terms sized by importance; chords are the automatically selected pairwise interactions, with width proportional to term importance. Sex-involving interactions are highlighted: these are the proxy pathways that make fairness-through-unawareness insufficient, and they are readable directly off the fitted model rather than inferred post hoc.}
\label{fig:interactions}
\end{figure}

At the pre-specified threshold the glass-box model detected 75.4\% of
women's prior infarctions against 89.0\% of men's --- a TPR gap of
0.128\,$\pm$\,0.008 across seeds, with XGBoost indistinguishable
(0.127\,$\pm$\,0.008). Under the common selection objective, per-group
thresholds closed the gap to 0.0125\,$\pm$\,0.0095, reweighing to
0.0090\,$\pm$\,0.0054 and shape repair with intercept equalization to
0.0104\,$\pm$\,0.0099 --- statistically matched arms (paired difference
$-0.0013$, 95\% bootstrap CI $-0.0049$ to $0.0022$) at essentially
identical specificity (Table~\ref{tab:fairness}; Supplementary
Figure~3). Deleting the sex terms
outright stalled at 0.043 while costing the most specificity: proxy
pathways carry most of the disparity, and the fitted model makes them
explicit, with sex entering the learned interaction structure through age,
general health and smoking status rather than in isolation
(Figure~\ref{fig:interactions}).

\FloatBarrier
 Closing the TPR gap closed the
false-positive-rate gap (0.141\,$\to$\,0.007) but widened the
positive-predictive-value gap (0.038\,$\to$\,0.070), making the
calibration/error-rate impossibility concrete. Baseline age-band gaps were
far larger (0.597) and closed only at heavy specificity cost; race and
ethnicity gaps of 0.243 closed to 0.082 at nearly retained specificity
(Supplementary Table~4).

\begin{table}[!htb]\centering
\caption{Fairness mitigation arms at the pre-specified screening
threshold. Shape-repair arms exist only for the glass-box model by
construction; XGBoost rows cover the model-agnostic arms.}
\label{tab:fairness}
{\footnotesize\setlength{\tabcolsep}{4pt}
\begin{tabular}{@{}llccccc@{}}
\toprule
Model & Arm & Sensitivity & Specificity & $\Delta$TPR (Sex) & $\Delta$FPR & $\Delta$PPV\\
\midrule
EBM & baseline (single $t$) & 0.848$\pm$0.008 & 0.679$\pm$0.007 & 0.128$\pm$0.008 & 0.141 & 0.038 \\
EBM & per-group thresholds$^{a}$ & 0.851$\pm$0.008 & 0.663$\pm$0.007 & 0.013$\pm$0.010 & 0.007 & 0.070 \\
EBM & reweighing (refit) & 0.848$\pm$0.006 & 0.662$\pm$0.004 & 0.009$\pm$0.005 & 0.004 & 0.069 \\
EBM & repair (attenuate Sex) & 0.851$\pm$0.008 & 0.670$\pm$0.005 & 0.041$\pm$0.009 & 0.042 & 0.062 \\
EBM & repair (zero Sex) & 0.850$\pm$0.009 & 0.648$\pm$0.006 & 0.043$\pm$0.009 & 0.064 & 0.083 \\
EBM & repair + equalise & 0.850$\pm$0.008 & 0.664$\pm$0.007 & 0.010$\pm$0.010 & 0.007 & 0.070 \\
\addlinespace
XGBoost & baseline (single $t$) & 0.847$\pm$0.005 & 0.679$\pm$0.005 & 0.127$\pm$0.008 & 0.143 & 0.038 \\
XGBoost & per-group thresholds$^{a}$ & 0.849$\pm$0.007 & 0.664$\pm$0.007 & 0.011$\pm$0.007 & 0.010 & 0.070 \\
XGBoost & reweighing (refit) & 0.848$\pm$0.008 & 0.661$\pm$0.005 & 0.012$\pm$0.007 & 0.006 & 0.069 \\
\addlinespace
\bottomrule
\end{tabular}

\smallskip
{\scriptsize Mean $\pm$ SD over five seeds; common selection objective
(validation TPR equalised at $\tau^{\ast}$). $^{a}$The
sensitivity-target and TPR-equalisation threshold variants coincide
numerically ($\tau^{\ast}\!\approx\!0.85$). Between-arm gap differences
cross zero under paired seed contrasts and a 1{,}000-draw stratified
bootstrap (repair vs.\ thresholds $-0.0013$ [$-0.0049$, $0.0022$]).
Shape-repair arms exist only for the glass-box model by construction;
effective per-group thresholds for the repair arm are reported in the
text. Age-band and race/ethnicity audits appear in Supplementary
Table~4.}
}
\end{table}
\FloatBarrier

\subsection*{Marginal conformal guarantees hide group-level failure}
\begin{figure}[!htb]\centering
\includegraphics[width=\linewidth]{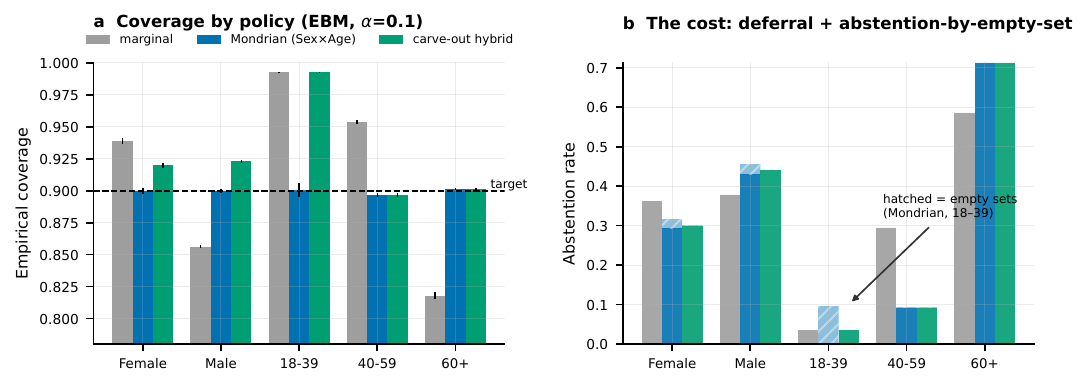}
\caption{\textbf{Marginal guarantees hide subgroup failure, and the repair has two currencies.} (\textbf{a}) Conformal coverage by policy for the glass-box model at $\alpha=0.1$ (mean $\pm$ SD over five seeds): marginal, Mondrian within sex$\times$age, and the carve-out hybrid. (\textbf{b}) The price: clinician deferral (solid) plus abstention through empty prediction sets (hatched), which only group-conditional calibration produces and only in the lowest-risk stratum.}
\label{fig:conformal}
\end{figure}

Split conformal prediction met its marginal guarantee overall but
redistributed it unevenly (Figure~\ref{fig:conformal}; Supplementary
Table~3): coverage was 0.938--0.940 for women against 0.856--0.858 for
men, and 0.992--0.993 for ages 18--39 against 0.818--0.819 for ages 60+,
so the group at highest cardiovascular risk received the weakest
guarantee. The pattern replicated across the glass-box model, XGBoost and
TabICL. Deferral tracked the same gradient, from 0.035 in adults 18--39 to
0.586 in adults 60+, whose singleton predictions were wrong 44\% of the
time. Mondrian calibration restored every stratum to 0.90 at two distinct
prices: honest deferral in the oldest stratum (0.586\,$\to$\,0.712), and,
in the youngest, empty prediction sets (9.5\%) that marginal calibration
never produces --- so the identity ``mean set size $=1+$ deferral'' holds
marginally but breaks under Mondrian. A hybrid retaining marginal
calibration for ages 18--39 and Mondrian elsewhere dominated both pure
policies under every utility weighting we examined.

\FloatBarrier

\subsection*{Post-hoc explanations are not glass-box explanations}
\begin{figure}[!htb]\centering
\includegraphics[width=\linewidth]{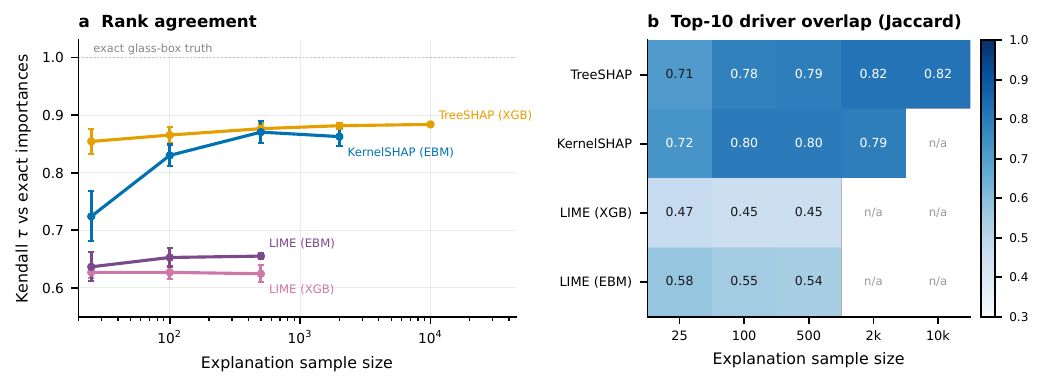}
\caption{\textbf{Post-hoc explanation is an estimate; glass-box explanation is an identity.} (\textbf{a}) Kendall $\tau$ between post-hoc attributions and the glass-box model's exact global importances versus explanation sample size; LIME applied to the glass-box model itself plateaus at $\tau\approx0.63$, acquitting the cross-model confound while confirming the estimator limitation. (\textbf{b}) Overlap of each explainer's ten strongest drivers with the exact top ten (Jaccard index); grey cells were not computed for tractability.}
\label{fig:faith}
\end{figure}

Against the glass-box model's exact global importances, TreeSHAP on tuned
XGBoost plateaued at Kendall $\tau=0.884$ even with 10\,000 explanation
samples (Figure~\ref{fig:faith}). More strikingly, model-agnostic
KernelSHAP applied to the glass-box model \emph{itself} recovered its own
knowable truth only gradually ($\tau=0.724$ at 25 samples, 0.870 at 500),
and LIME never converged at any budget ($\tau\approx0.626$; top-10 Jaccard
0.45, misidentifying roughly half of the ten strongest drivers).

\FloatBarrier

\subsection*{Frozen models and thresholds transport to the next survey year}
Applied without refitting or re-thresholding to all 430\,755 respondents of
2023, every T1-portable model lost between 0.0009 and 0.0016 AUROC relative
to its internal estimate (Table~\ref{tab:temporal}), calibration slopes
stayed within 0.91--1.19, and the frozen 2022 thresholds delivered 2023
sensitivities of 0.855--0.862 against the 0.85 target. Glass-box shape
functions refitted independently on 2023 reproduced the 2022 risk curves
within bagged uncertainty bands (Supplementary Figure~4).

\begin{table}[!htb]\centering
\caption{Temporal external validation: all 2022 models and 2022
thresholds applied frozen to the full 2023 cycle.}
\label{tab:temporal}
{\footnotesize\setlength{\tabcolsep}{5pt}
\begin{tabular}{@{}lccccc@{}}
\toprule
Model & AUROC & AUPRC & Cal.\ slope & Sens.@screen & Spec.@screen\\
\midrule
LogReg & 0.8378 & 0.2465 & 0.928 & 0.861 & 0.651 \\
LogReg (splines) & 0.8375 & 0.2469 & 0.915 & 0.862 & 0.648 \\
Random Forest & 0.8373 & 0.2405 & 1.191 & 0.856 & 0.659 \\
EBM & 0.8395 & 0.2511 & 0.958 & 0.859 & 0.656 \\
XGBoost & 0.8400 & 0.2511 & 1.001 & 0.858 & 0.659 \\
LightGBM & 0.8399 & 0.2506 & 1.025 & 0.856 & 0.660 \\
CatBoost & 0.8410 & 0.2527 & 1.022 & 0.858 & 0.661 \\
MLP & 0.8401 & 0.2505 & 1.039 & 0.861 & 0.655 \\
TabPFN v2$^{\dagger}$ & 0.8373 & 0.2456 & 0.928 & 0.855 & 0.658 \\
TabICL & 0.8411 & 0.2524 & 0.926 & 0.857 & 0.662 \\
\bottomrule
\end{tabular}

\smallskip
{\scriptsize All 2022 models and 2022 thresholds applied frozen to the
full 2023 cycle ($n{=}430{,}755$); values are means over the five seeded frozen models, at four
decimals. $^{\dagger}$Subsampled-context protocol.}
}
\end{table}
\FloatBarrier

\subsection*{Sensitivity analyses and the survey-weighted population}
SMOTE-NC oversampling, standard practice in this literature, reduced T1
AUROC from 0.843 to 0.809 while inflating training time
$\sim$150-fold.\cite{chawla2002smote} Complete-case analysis cost 0.0023
AUROC in addition to discarding 42.9\% of respondents. Re-evaluating every
frozen model under survey design weights raised discrimination almost
uniformly, by $+0.020$ to $+0.022$ AUROC for all ten models, leaving the
ranking statistically unchanged (Figure~\ref{fig:leakage}d). Testing rather
than assuming where a foundation-model edge might lie, we compared TabICL
with CatBoost within design-weight quintiles: differences were null in
quintiles 1--4, and in the highest-weight quintile --- the most
under-sampled respondents --- the sign ran against the foundation model
(CatBoost better by 0.0029, 95\% CI 0.0009--0.0048). Weight-informed
training did not improve weighted-population discrimination; the foundation
models cannot ingest per-instance weights in their zero-shot interface.

\section*{DISCUSSION}

\subsection*{Principal findings}
The headroom celebrated in this literature is leakage, not learning. Every
model we trained, from logistic regression to a 265\,240-row in-context
foundation model, reproduced the $\sim$0.89 regime on the full feature set
and surrendered roughly 0.05 AUROC when two post-diagnostic markers were
removed. The near-uniformity of that decline is the finding: it gives
quantitative, dose--response form to the leakage taxonomies of Davis et al.\cite{davis2023leakage} and to the
diagnosis-code case reports of Ramadan et al.,\cite{ramadan2025leakage} and it generalises the
single-pipeline correction of Eltawil et al.\cite{eltawil2026smote} to ten
models and five tiers. For the many public tools built on redistributed
derivatives of this survey, the implication is uncomfortable: a model that
knows a respondent has angina is not screening for myocardial infarction,
it is transcribing a record back to itself.

Transparency cost nothing measurable. At the reduced-leakage tier the
glass-box model was statistically non-inferior to every alternative within
a pre-specified margin, extending the intelligible-models result of Caruana
et al.\cite{caruana2015intelligible} to the foundation-model era
while scoring the cohort four orders of magnitude faster than TabICL at
equal accuracy. This is not a verdict against foundation models: TabICL
matched the best boosted trees with zero tuning, and both foundation models
arrived natively calibrated where every class-weighted classical model
required isotonic repair --- an under-appreciated consequence of imbalance
handling, and a caution for any study reporting class-weighted
probabilities as risks. But the cost asymmetry, the weakness of the
subsampled-context variant, and the shrinkage of any foundation-model edge
under survey weighting together suggest that, for population screening on
tabular surveys, in-context scale currently buys parity rather than
superiority.

Fairness must be engineered at the operating point. Reporting a single
validation-selected threshold converts this literature's threshold-free
disparity claims into an explicit one: one in four true cases among women
is missed against one in nine among men. That the gap was essentially
identical for the glass-box model and XGBoost shows it is a property of the
data-plus-threshold system, not of any model's opacity. Shape repair with
intercept equalization matched conventional mitigation at matched
specificity while remaining what the alternatives, including
fairness-aware interpretable modelling,\cite{faim2024} are not --- a reversible,
machine-readable, governable edit of the model itself, with its per-group
effective thresholds disclosed rather than implicit. This converts the
interactive editing paradigm of Wang et al.\cite{wang2022gamchanger} into
an auditable fairness protocol, and the same additive structure makes visible why
fairness-through-unawareness fails.

Uncertainty guarantees are only as good as their conditioning. Marginal
conformal prediction met its 90\% guarantee on average while delivering
86\% coverage to men and 82\% to adults over sixty --- precisely the
stratum at highest risk --- replicating across three model families the
group-coverage failures anticipated by the conditional-conformal
literature\cite{gibbs2023conditional} and motivating fairness-aware
conformal auditing at the point of care.\cite{sun2025fairicp} We add that
group-conditional repair in very-low-risk strata can also abstain through
empty prediction sets, a cost invisible to marginal
accounting.\cite{kwon2026conformal} Finally, the faithfulness analysis
cautions the field's default explanation practice: even explaining the
glass-box model itself, KernelSHAP needed hundreds of samples and LIME
never converged. Where stakes justify explanation, they justify models
whose explanations are identities rather than estimates.

\subsection*{Limitations}
The outcome is self-reported, ever-diagnosed myocardial infarction in a
cross-sectional survey: it identifies prevalent, surviving,
diagnosis-aware cases, so the task is screening triage rather than incident
risk prediction, and fatal or undiagnosed events are structurally
invisible. Self-report introduces misclassification on outcome and
predictors alike. Our tiers operationalise one principled leakage taxonomy;
other boundaries are defensible, and T1 removes only \emph{direct}
post-diagnostic markers, so residual conditioning-on-care signal remains in
variables such as checkup recency. Survey-weighted results are
point-estimate sensitivity analyses; full design-based variance was out of
scope. Both validation years come from the same surveillance system and
country, so transportability to clinical registries or other health systems
is untested. Foundation models evolve quickly, and successors may behave
differently. Fairness was audited on measured attributes only, and
gap closure at one threshold does not certify equity of downstream care.
Conformal validity assumes exchangeability that drift can break; our 2023
results show empirical robustness for one year, not a guarantee.

\section*{CONCLUSION}

On the most heavily reused public cardiovascular survey data, two
post-diagnostic features --- not model sophistication --- explain the
performance this literature celebrates. Once they are removed, a
transparent, editable, sub-second glass-box model meets tuned gradient
boosting and full-context tabular foundation models head-on, and its
structure supports what deployment actually requires: explicit operating
points, auditable fairness repair that matches conventional mitigation
while remaining inspectable, group-conditional uncertainty with honest
deferral, exact rather than estimated explanations, and thresholds that
transport across survey years. The binding constraint on this field is
evaluation practice, not model capacity.

\section*{Data availability}
This study analysed third-party data that are publicly available without
restriction. The 2022 and 2023 Behavioral Risk Factor Surveillance System
Landline and Cellular Telephone files are distributed by the Centers for
Disease Control and Prevention and are cited in the reference
list.\cite{cdc2022brfss,cdc2023brfss} No raw data are redistributed here.
Derived cohort construction logs and every per-cell result reported in this
article accompany the code release.

\section*{Code availability}
The complete configuration-driven pipeline --- cohort construction, models,
audits, statistical tests, tables and figures --- is openly available at
\url{https://github.com/raadbintareaf/cvd-glassbox-screening} and archived at
\url{https://doi.org/10.5281/zenodo.21918761}.

\section*{Funding}
Supported by Abu Dhabi University's Office of Research and Sponsored Programs
(grant number 19300893).

\section*{Author contributions}
Following CRediT: \textbf{R.B.T.} --- conceptualization, methodology, software,
formal analysis, data curation, visualization, writing (original draft).
\textbf{M.A.R.} and \textbf{S.L.} --- supervision, validation, writing (review
and editing). \textbf{S.E.} --- validation (clinical plausibility of learned
risk functions), writing (review and editing). \textbf{C.S.} --- investigation
(literature synthesis), writing (review and editing).

\section*{Competing interests}
The authors declare no competing interests.

\bibliographystyle{unsrtnat}
\bibliography{refs}

@article{hollmann2025tabpfn,
  author={Hollmann, Noah and others},
  title={{Accurate predictions on small data with a tabular foundation model}},
  journal={Nature},
  year={2025},
  volume={637},
  pages={319-326},
  doi={10.1038/s41586-024-08328-6}
}

@inproceedings{wang2022gamchanger,
  author={Wang, Zijie J. and others},
  title={{Interpretability, Then What? Editing Machine Learning Models to Reflect Human Knowledge and Values}},
  booktitle={Proceedings of the 28th ACM SIGKDD Conference on Knowledge Discovery and Data Mining},
  year={2022},
  pages={4132-4142},
  doi={10.1145/3534678.3539074}
}

@article{davis2023leakage,
  author={Davis, Sharon E and Matheny, Michael E and Balu, Suresh and Sendak, Mark P},
  title={{A framework for understanding label leakage in machine learning for health care}},
  journal={Journal of the American Medical Informatics Association},
  year={2023},
  volume={31},
  pages={274-280},
  doi={10.1093/jamia/ocad178}
}

@article{ramadan2025leakage,
  author={Ramadan, Bashar and Liu, Ming-Chieh and Burkhart, Michael C. and Parker, William F. and Beaulieu-Jones, Brett K.},
  title={{Diagnostic Codes in AI Prediction Models and Label Leakage of Same-Admission Clinical Outcomes}},
  journal={JAMA Network Open},
  year={2025},
  volume={8},
  pages={e2550454},
  doi={10.1001/jamanetworkopen.2025.50454}
}

@article{eltawil2026smote,
  author={Eltawil, Mohamed and Byham-Gray, Laura and Jia, Yuane and Mistry, Neil and Parrott, James and Gohel, Suril},
  title={{Comment on Iacobescu et al. Evaluating Binary Classifiers for Cardiovascular Disease Prediction: Enhancing Early Diagnostic Capabilities. J. Cardiovasc. Dev. Dis. 2024, 11, 396}},
  journal={Journal of Cardiovascular Development and Disease},
  year={2026},
  volume={13},
  pages={46},
  doi={10.3390/jcdd13010046}
}

@article{gibbs2023conditional,
  author={Gibbs, Isaac and Cherian, John J and Candès, Emmanuel J},
  title={{Conformal prediction with conditional guarantees}},
  journal={Journal of the Royal Statistical Society Series B: Statistical Methodology},
  year={2025},
  volume={87},
  pages={1100-1126},
  doi={10.1093/jrsssb/qkaf008}
}

@article{kwon2026conformal,
  author={Kwon, Hyun and Kim, Dae-Jin},
  title={{Conformal selective prediction with cost aware deferral for safe clinical triage under distribution shift}},
  journal={Scientific Reports},
  year={2026},
  volume={16},
  doi={10.1038/s41598-026-40637-w}
}

@article{sun2025fairicp,
  author={Sun, Xiaotan and others},
  title={{FairICP: identifying biases and increasing transparency at the point of care in post-implementation clinical decision support using inductive conformal prediction}},
  journal={Journal of the American Medical Informatics Association},
  year={2025},
  volume={32},
  pages={1299-1309},
  doi={10.1093/jamia/ocaf095}
}

@article{faim2024,
  author={Liu, Mingxuan and others},
  title={{FAIM: Fairness-aware interpretable modeling for trustworthy machine learning in healthcare}},
  journal={Patterns},
  year={2024},
  volume={5},
  pages={101059},
  doi={10.1016/j.patter.2024.101059}
}

@article{collins2024tripod,
  author={Collins, Gary S and others},
  title={{TRIPOD+AI statement: updated guidance for reporting clinical prediction models that use regression or machine learning methods}},
  journal={BMJ},
  year={2024},
  pages={e078378},
  doi={10.1136/bmj-2023-078378}
}

@inproceedings{caruana2015intelligible,
  author={Caruana, Rich and Lou, Yin and Gehrke, Johannes and Koch, Paul and Sturm, Marc and Elhadad, Noemie},
  title={{Intelligible Models for HealthCare}},
  booktitle={Proceedings of the 21th ACM SIGKDD International Conference on Knowledge Discovery and Data Mining},
  year={2015},
  pages={1721-1730},
  doi={10.1145/2783258.2788613}
}

@inproceedings{lou2013ga2m,
  author={Lou, Yin and Caruana, Rich and Gehrke, Johannes and Hooker, Giles},
  title={{Accurate intelligible models with pairwise interactions}},
  booktitle={Proceedings of the 19th ACM SIGKDD international conference on Knowledge discovery and data mining},
  year={2013},
  pages={623-631},
  doi={10.1145/2487575.2487579}
}

@inproceedings{ribeiro2016lime,
  author={Ribeiro, Marco Tulio and Singh, Sameer and Guestrin, Carlos},
  title={{"Why Should I Trust You?"}},
  booktitle={Proceedings of the 22nd ACM SIGKDD International Conference on Knowledge Discovery and Data Mining},
  year={2016},
  pages={1135-1144},
  doi={10.1145/2939672.2939778}
}

@inproceedings{chen2016xgboost,
  author={Chen, Tianqi and Guestrin, Carlos},
  title={{XGBoost}},
  booktitle={Proceedings of the 22nd ACM SIGKDD International Conference on Knowledge Discovery and Data Mining},
  year={2016},
  pages={785-794},
  doi={10.1145/2939672.2939785}
}

@article{breiman2001rf,
  author={Breiman, Leo},
  title={{Random Forests}},
  journal={Machine Learning},
  year={2001},
  volume={45},
  pages={5-32},
  doi={10.1023/A:1010933404324}
}

@article{delong1988,
  author={DeLong, Elizabeth R. and DeLong, David M. and Clarke-Pearson, Daniel L.},
  title={{Comparing the Areas under Two or More Correlated Receiver Operating Characteristic Curves: A Nonparametric Approach}},
  journal={Biometrics},
  year={1988},
  volume={44},
  pages={837},
  doi={10.2307/2531595}
}

@book{vovk2005conformal,
  author={Vovk, Vladimir and Gammerman, Alexander and Shafer, Glenn},
  title={Algorithmic Learning in a Random World},
  publisher={Springer},
  address={New York},
  year={2005},
  doi={10.1007/b106715}
}

@article{kamiran2012reweighing,
  author={Kamiran, Faisal and Calders, Toon},
  title={{Data preprocessing techniques for classification without discrimination}},
  journal={Knowledge and Information Systems},
  year={2012},
  volume={33},
  pages={1-33},
  doi={10.1007/s10115-011-0463-8}
}

@article{vickers2006dca,
  author={Vickers, Andrew J. and Elkin, Elena B.},
  title={{Decision Curve Analysis: A Novel Method for Evaluating Prediction Models}},
  journal={Medical Decision Making},
  year={2006},
  volume={26},
  pages={565-574},
  doi={10.1177/0272989X06295361}
}

@inproceedings{akiba2019optuna,
  author={Akiba, Takuya and Sano, Shotaro and Yanase, Toshihiko and Ohta, Takeru and Koyama, Masanori},
  title={{Optuna}},
  booktitle={Proceedings of the 25th ACM SIGKDD International Conference on Knowledge Discovery \& Data Mining},
  year={2019},
  pages={2623-2631},
  doi={10.1145/3292500.3330701}
}

@article{chawla2002smote,
  author={Chawla, N. V. and Bowyer, K. W. and Hall, L. O. and Kegelmeyer, W. P.},
  title={{SMOTE: Synthetic Minority Over-sampling Technique}},
  journal={Journal of Artificial Intelligence Research},
  year={2002},
  volume={16},
  pages={321-357},
  doi={10.1613/jair.953}
}

@inproceedings{zadrozny2002isotonic,
  author={Zadrozny, Bianca and Elkan, Charles},
  title={{Transforming classifier scores into accurate multiclass probability estimates}},
  booktitle={Proceedings of the eighth ACM SIGKDD international conference on Knowledge discovery and data mining},
  year={2002},
  pages={694-699},
  doi={10.1145/775047.775151}
}

@inproceedings{lundberg2017shap,
  author={Lundberg, Scott M. and Lee, Su-In},
  title={{A Unified Approach to Interpreting Model Predictions}},
  booktitle={Advances in Neural Information Processing Systems 30},
  year={2017}, note={arXiv:1705.07874}
}

@inproceedings{ke2017lightgbm,
  author={Ke, Guolin and others},
  title={{LightGBM: A Highly Efficient Gradient Boosting Decision Tree}},
  booktitle={Advances in Neural Information Processing Systems 30},
  year={2017},
  pages={3146-3154}
}

@inproceedings{prokhorenkova2018catboost,
  author={Prokhorenkova, Liudmila and others},
  title={{CatBoost: unbiased boosting with categorical features}},
  booktitle={Advances in Neural Information Processing Systems 31},
  year={2018}, note={arXiv:1706.09516}
}

@inproceedings{qu2025tabicl,
  author={Qu, Jingang and others},
  title={{TabICL: A Tabular Foundation Model for In-Context Learning on Large Data}},
  booktitle={Proceedings of the 42nd International Conference on Machine Learning},
  year={2025}, note={arXiv:2502.05564}
}

@misc{nori2019interpretml,
  author={Nori, Harsha and Jenkins, Samuel and Koch, Paul and Caruana, Rich},
  title={{InterpretML: A Unified Framework for Machine Learning Interpretability}},
  year={2019}, note={arXiv:1909.09223}
}

@article{holm1979,
  author={Holm, Sture},
  title={{A Simple Sequentially Rejective Multiple Test Procedure}},
  journal={Scandinavian Journal of Statistics},
  year={1979}, volume={6}, pages={65--70}
}

@misc{cdc2022brfss,
  author={{Centers for Disease Control and Prevention}},
  title={{Behavioral Risk Factor Surveillance System: 2022 Survey Data and Documentation (LLCP2022)}},
  year={2023}, howpublished={\url{https://www.cdc.gov/brfss/annual_data/annual_2022.html}}
}

@misc{cdc2023brfss,
  author={{Centers for Disease Control and Prevention}},
  title={{Behavioral Risk Factor Surveillance System: 2023 Survey Data and Documentation (LLCP2023)}},
  year={2024}, howpublished={\url{https://www.cdc.gov/brfss/annual_data/annual_2023.html}}
}

\clearpage
\appendix
\section*{Supplementary Information}
\addcontentsline{toc}{section}{Supplementary Information}
\refstepcounter{section}

\noindent This appendix contains the material referred to in the main text as
Supplementary Figures 1--4, Supplementary Tables 1--4, Supplementary Methods
and the Supplementary Note. It is identical to the Supplementary Information
filed with the journal, and is bound here so that the preprint is a single
self-contained document.

\subsection*{Supplementary Figures}
\suppfig{Supplementary Fig.~1 — Cohort construction}
\begin{figure}[h!]\centering
\includegraphics[width=.92\linewidth]{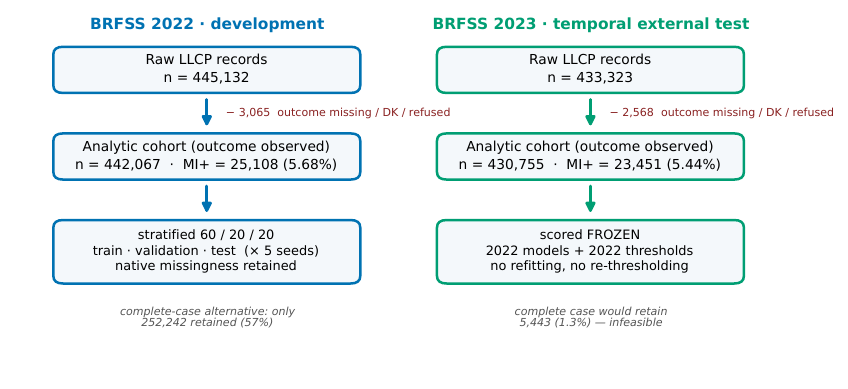}
\caption*{\textbf{Supplementary Fig.~1 $|$ Cohort construction.} Record
flow for the 2022 development and 2023 temporal-validation cycles;
complete-case counts shown for reference only (primary analyses retain
native missingness).}\end{figure}

\clearpage
\suppfig{Supplementary Fig.~2 — Cohort geography and demography}
\begin{figure}[h!]\centering
\includegraphics[width=\linewidth]{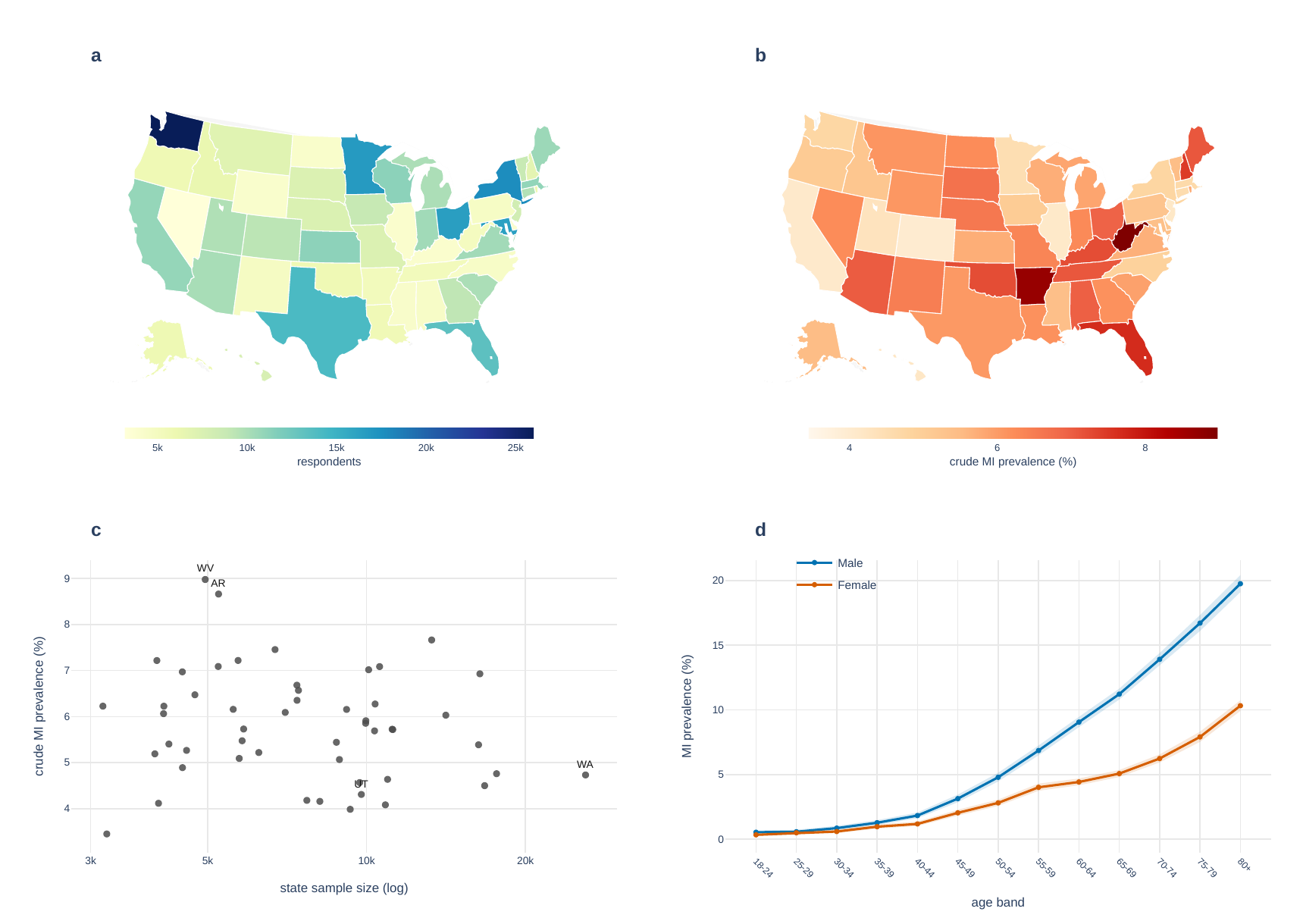}
\caption*{\textbf{Supplementary Fig.~2 $|$ Cohort geography and
demography (2022).} (\textbf{a}) Respondents per state and (\textbf{b})
crude self-reported MI prevalence (unweighted; territories GU/PR/VI,
$n{=}9{,}258$, not mapped). (\textbf{c}) State prevalence versus sample
size. (\textbf{d}) Age--sex prevalence gradient with 95\% Wilson
intervals: male excess emerges in midlife and persists to 80+.}
\end{figure}

\clearpage
\suppfig{Supplementary Fig.~3 — Equity–specificity plane of the mitigation arms}
\begin{center}\includegraphics[width=.78\linewidth]{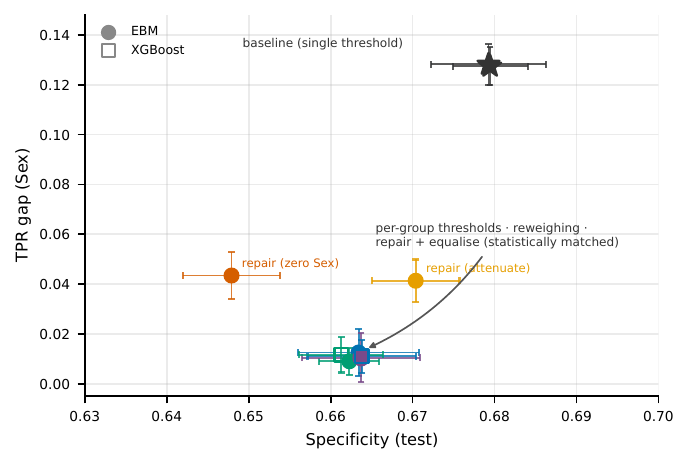}\end{center}
\noindent\textbf{Supplementary Fig.~3 $|$ Equity--specificity plane of the
mitigation arms (Sex).} Mean $\pm$ SD over five seeds under the common
selection objective; filled circles the glass-box model, open squares
XGBoost, star the shared single-threshold baseline. Per-group thresholds,
reweighing, and shape repair with intercept equalisation occupy the same
corner of the plane---statistically matched gap closure at matched
specificity---while partial and unawareness repairs stall mid-plane.
Numerical values, including the false-positive-rate and
positive-predictive-value gaps this projection does not show, are given in
Table~\ref{tab:fairness}.

\clearpage
\suppfig{Supplementary Fig.~4 — Learned risk shapes and their replication across survey years}
\begin{center}\includegraphics[width=.95\linewidth]{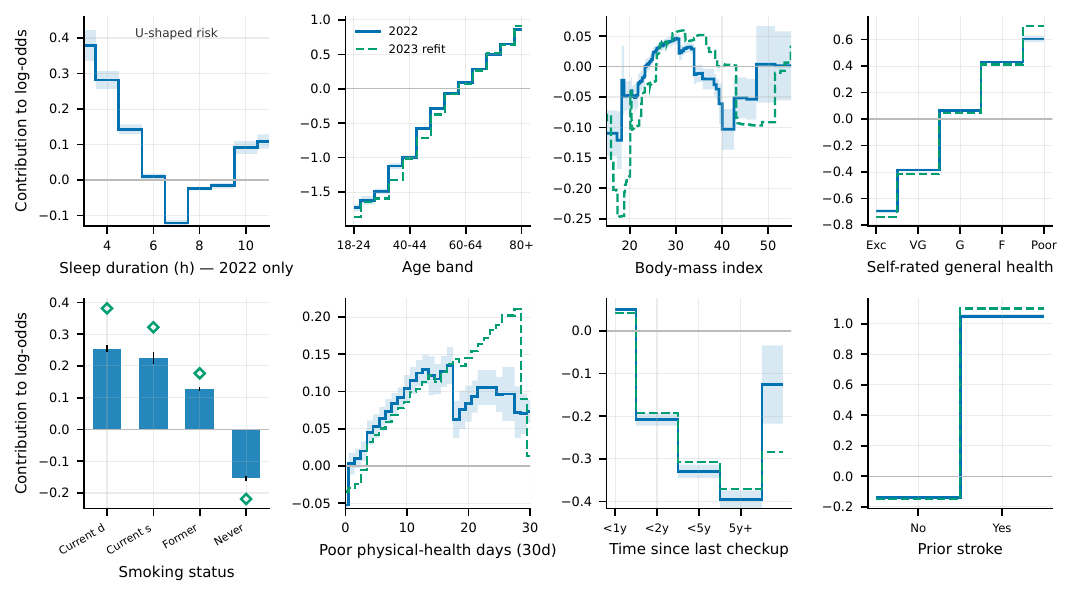}\end{center}
\noindent\textbf{Supplementary Fig.~4 $|$ Learned risk shapes and their
replication across survey years.} Glass-box term contributions for
continuous, ordinal and categorical predictors: 2022 (with bagged
uncertainty) versus an independent 2023 refit on the year-portable tier.
Sleep duration was removed from the 2023 core questionnaire and is
therefore shown for 2022 only.

\clearpage
\subsection*{Supplementary Tables}
\suppfig{Supplementary Table 1 — Variable map and tier membership}
{\scriptsize\begin{longtable}{@{}lllccc@{}}
\caption{Analytic variables, raw BRFSS items, and tier membership.}\\
\toprule
Analytic name & 2022 item & 2023 item & T1 & T1-port. & T2\\
\midrule
\endfirsthead
\toprule
Analytic name & 2022 item & 2023 item & T1 & T1-port. & T2\\
\midrule
\endhead
AgeCategory & -- & -- & \checkmark & \checkmark & \checkmark \\
AlcoholDrinkers & -- & -- & \checkmark & \checkmark & \checkmark \\
BMI & -- & -- & \checkmark & \checkmark & \checkmark \\
BlindVisionDiff & -- & -- & \checkmark & \checkmark & -- \\
ChestScan & -- & -- & -- & -- & -- \\
CovidPositive & -- & -- & \checkmark & \checkmark & -- \\
DeafHardHearing & -- & -- & \checkmark & \checkmark & -- \\
DiabetesStatus & -- & -- & \checkmark & \checkmark & -- \\
DiffConcentrating & -- & -- & \checkmark & \checkmark & -- \\
DiffDressing & -- & -- & \checkmark & \checkmark & -- \\
DiffErrands & -- & -- & \checkmark & \checkmark & -- \\
DiffWalking & -- & -- & \checkmark & \checkmark & -- \\
ECigaretteUsage & -- & -- & \checkmark & \checkmark & \checkmark \\
FluVaxLast12 & -- & -- & \checkmark & \checkmark & -- \\
GeneralHealth & -- & -- & \checkmark & \checkmark & \checkmark \\
HIVTesting & -- & -- & \checkmark & \checkmark & -- \\
HadAngina & -- & -- & -- & -- & -- \\
HadArthritis & -- & -- & \checkmark & \checkmark & -- \\
HadAsthma & -- & -- & \checkmark & \checkmark & -- \\
HadCOPD & -- & -- & \checkmark & \checkmark & -- \\
HadDepression & -- & -- & \checkmark & \checkmark & -- \\
HadKidneyDisease & -- & -- & \checkmark & \checkmark & -- \\
HadOtherCancer & -- & -- & \checkmark & \checkmark & -- \\
HadSkinCancer & -- & -- & \checkmark & \checkmark & -- \\
HadStroke & -- & -- & \checkmark & \checkmark & -- \\
HeightMeters & -- & -- & \checkmark & \checkmark & \checkmark \\
LastCheckupTime & -- & -- & \checkmark & \checkmark & -- \\
MentalHealthDays & -- & -- & \checkmark & \checkmark & \checkmark \\
PhysicalActivities & -- & -- & \checkmark & \checkmark & \checkmark \\
PhysicalHealthDays & -- & -- & \checkmark & \checkmark & \checkmark \\
PneumoVaxEver & -- & -- & \checkmark & \checkmark & -- \\
RaceEthnicity & -- & -- & \checkmark & \checkmark & \checkmark \\
RemovedTeeth & -- & -- & \checkmark & -- & -- \\
Sex & -- & -- & \checkmark & \checkmark & \checkmark \\
SleepHours & -- & -- & \checkmark & -- & \checkmark \\
SmokerStatus & -- & -- & \checkmark & \checkmark & \checkmark \\
State & -- & -- & \checkmark & \checkmark & \checkmark \\
TetanusLast10 & -- & -- & \checkmark & -- & -- \\
WeightKilograms & -- & -- & \checkmark & \checkmark & \checkmark \\
\bottomrule
\end{longtable}
}

\clearpage
\suppfig{Supplementary Table 2 — Per-tier feature lists}
\noindent\textbf{Supplementary Table 2 $|$ Per-tier feature lists.}\\[4pt]
{\small\begin{description}
\item[T0 (39 features)] AgeCategory, AlcoholDrinkers, BMI, BlindVisionDiff, ChestScan, CovidPositive, DeafHardHearing, DiabetesStatus, DiffConcentrating, DiffDressing, DiffErrands, DiffWalking, ECigaretteUsage, FluVaxLast12, GeneralHealth, HIVTesting, HadAngina, HadArthritis, HadAsthma, HadCOPD, HadDepression, HadKidneyDisease, HadOtherCancer, HadSkinCancer, HadStroke, HeightMeters, LastCheckupTime, MentalHealthDays, PhysicalActivities, PhysicalHealthDays, PneumoVaxEver, RaceEthnicity, RemovedTeeth, Sex, SleepHours, SmokerStatus, State, TetanusLast10, WeightKilograms.
\item[T1 (37 features)] AgeCategory, AlcoholDrinkers, BMI, BlindVisionDiff, CovidPositive, DeafHardHearing, DiabetesStatus, DiffConcentrating, DiffDressing, DiffErrands, DiffWalking, ECigaretteUsage, FluVaxLast12, GeneralHealth, HIVTesting, HadArthritis, HadAsthma, HadCOPD, HadDepression, HadKidneyDisease, HadOtherCancer, HadSkinCancer, HadStroke, HeightMeters, LastCheckupTime, MentalHealthDays, PhysicalActivities, PhysicalHealthDays, PneumoVaxEver, RaceEthnicity, RemovedTeeth, Sex, SleepHours, SmokerStatus, State, TetanusLast10, WeightKilograms.
\item[T1-portable (34 features)] AgeCategory, AlcoholDrinkers, BMI, BlindVisionDiff, CovidPositive, DeafHardHearing, DiabetesStatus, DiffConcentrating, DiffDressing, DiffErrands, DiffWalking, ECigaretteUsage, FluVaxLast12, GeneralHealth, HIVTesting, HadArthritis, HadAsthma, HadCOPD, HadDepression, HadKidneyDisease, HadOtherCancer, HadSkinCancer, HadStroke, HeightMeters, LastCheckupTime, MentalHealthDays, PhysicalActivities, PhysicalHealthDays, PneumoVaxEver, RaceEthnicity, Sex, SmokerStatus, State, WeightKilograms.
\item[T2 (15 features)] AgeCategory, AlcoholDrinkers, BMI, ECigaretteUsage, GeneralHealth, HeightMeters, MentalHealthDays, PhysicalActivities, PhysicalHealthDays, RaceEthnicity, Sex, SleepHours, SmokerStatus, State, WeightKilograms.
\end{description}
}

\vspace{10pt}
\suppfig{Supplementary Table 3 — Conformal coverage, deferral and empty sets}
\noindent\textbf{Supplementary Table 3 $|$ Conformal coverage, deferral
and empty-set rates across the full model$\times$method grid.}\\[2pt]
{\small
\setlength{\tabcolsep}{3.5pt}
\begin{tabular}{@{}llccccc@{}}
\toprule
Model & Method & Female & Male & 18--39 & 40--59 & 60+\\
\midrule
\multicolumn{7}{@{}l}{\emph{Empirical coverage} (target 0.90)}\\
EBM & marginal & 0.9391 $\pm$ 0.0021 & 0.8561 $\pm$ 0.0012 & 0.9926 $\pm$ 0.0004 & 0.9538 $\pm$ 0.0016 & 0.8179 $\pm$ 0.0030 \\
EBM & Mondrian (Sex) & 0.901 $\pm$ 0.001 & 0.900 $\pm$ 0.002 & 0.992 $\pm$ 0.000 & 0.951 $\pm$ 0.001 & 0.820 $\pm$ 0.001 \\
EBM & Mondrian (Sex$\times$Age) & 0.8999 $\pm$ 0.0023 & 0.8999 $\pm$ 0.0014 & 0.9005 $\pm$ 0.0055 & 0.8967 $\pm$ 0.0013 & 0.9013 $\pm$ 0.0011 \\
EBM & carve-out hybrid$^{a}$ & 0.9199 & 0.9230 & 0.9926 & 0.8967 & 0.9013 \\
\addlinespace
XGBoost & marginal & 0.940 $\pm$ 0.001 & 0.856 $\pm$ 0.001 & 0.993 $\pm$ 0.001 & 0.953 $\pm$ 0.001 & 0.818 $\pm$ 0.001 \\
XGBoost & Mondrian (Sex) & 0.901 $\pm$ 0.002 & 0.901 $\pm$ 0.002 & 0.993 $\pm$ 0.001 & 0.951 $\pm$ 0.001 & 0.821 $\pm$ 0.002 \\
XGBoost & Mondrian (Sex$\times$Age) & 0.900 $\pm$ 0.002 & 0.900 $\pm$ 0.002 & 0.901 $\pm$ 0.005 & 0.898 $\pm$ 0.001 & 0.901 $\pm$ 0.001 \\
\addlinespace
TabICL & marginal & 0.938 $\pm$ 0.003 & 0.858 $\pm$ 0.001 & 0.992 $\pm$ 0.000 & 0.953 $\pm$ 0.001 & 0.818 $\pm$ 0.003 \\
TabICL & Mondrian (Sex) & 0.901 $\pm$ 0.001 & 0.900 $\pm$ 0.001 & 0.992 $\pm$ 0.000 & 0.951 $\pm$ 0.002 & 0.820 $\pm$ 0.002 \\
TabICL & Mondrian (Sex$\times$Age) & 0.901 $\pm$ 0.002 & 0.901 $\pm$ 0.002 & 0.902 $\pm$ 0.003 & 0.898 $\pm$ 0.002 & 0.902 $\pm$ 0.003 \\
\midrule
\multicolumn{7}{@{}l}{\emph{Deferral rate} (EBM; fraction of ambiguous $\{0,1\}$ sets)}\\
EBM & marginal & 0.362 & 0.378 & 0.035 & 0.295 & 0.586 \\
EBM & Mondrian (Sex$\times$Age) & 0.294 & 0.430 & 0.000 & 0.092 & 0.712 \\
EBM & carve-out hybrid$^{a}$ & 0.300 & 0.441 & 0.035 & 0.092 & 0.712 \\
\midrule
\multicolumn{7}{@{}l}{\emph{Empty prediction sets} (EBM, Mondrian; marginal produces none)}\\
EBM & Mondrian (Sex$\times$Age) & 2.1\% & 2.4\% & 9.5\% & 0.0\% & 0.0\% \\
\bottomrule
\end{tabular}

\smallskip
{\scriptsize Split conformal at $\alpha=0.1$; mean $\pm$ SD over five
seeds; target coverage 0.90 per group. $^{a}$The carve-out hybrid keeps
marginal calibration for ages 18--39 and Mondrian (Sex$\times$Age)
elsewhere. The accounting identity ``average set size $=1+$ deferral''
holds for the marginal method (zero empty sets) and breaks under
Mondrian in the lowest-risk stratum. Respondents with missing age
($n{=}1{,}795$) are excluded from the age strata. Four decimals are
shown for the EBM rows underlying the main-text claims.}
}

\clearpage
\suppfig{Supplementary Table 4 — Age-band and race/ethnicity fairness audits}
\noindent\textbf{Supplementary Table 4 $|$ Fairness audits beyond
sex.}\\[2pt]
{\small\begin{tabular}{@{}lllcccc@{}}
\toprule
Attribute & Model & Arm & $\Delta$TPR & $\Delta$PPV & Sens. & Spec.\\
\midrule
Age band & EBM & baseline (single $t$) & 0.597 & 0.024 & 0.838 & 0.684 \\
Age band & EBM & per-group thresholds & 0.073 & 0.150 & 0.841 & 0.559 \\
\addlinespace
Age band & XGBoost & baseline (single $t$) & 0.591 & 0.026 & 0.840 & 0.685 \\
Age band & XGBoost & per-group thresholds & 0.087 & 0.149 & 0.834 & 0.569 \\
\addlinespace
Race/ethnicity & EBM & baseline (single $t$) & 0.243 & 0.083 & 0.838 & 0.684 \\
Race/ethnicity & EBM & per-group thresholds & 0.082 & 0.159 & 0.839 & 0.677 \\
\addlinespace
Race/ethnicity & XGBoost & baseline (single $t$) & 0.184 & 0.088 & 0.840 & 0.685 \\
Race/ethnicity & XGBoost & per-group thresholds & 0.082 & 0.166 & 0.843 & 0.675 \\
\addlinespace
\bottomrule
\end{tabular}

\smallskip
{\scriptsize Seed-0 audits at the common objective; the
sensitivity-target threshold variant coincides with the
TPR-equalisation variant shown. Shape-repair arms target Sex terms and
are therefore not defined for these attributes. Age-conditioned
thresholds close the raw gap only at heavy specificity and PPV cost
(main text).}
}

\clearpage
\subsection*{Supplementary Methods}

\subsubsection*{Splits, tuning and the pre-specification regime}
Each tier was run with five random seeds under stratified 60/20/20
train/validation/test splits, stratified on the outcome and drawn
independently per seed. Hyperparameters were tuned with Optuna
(15--30 trials per model per tier) against validation AUROC on the seed-0
split only; the selected configurations were then frozen and reused for all
seeds, so seed-level standard deviations reflect split and fitting
variability rather than tuning variability. The two tabular foundation
models were deliberately not tuned, since zero-shot operation is their
claimed mode. Class imbalance was handled by balanced class weighting for
the classical models. No test-partition quantity informed any modelling,
tuning, thresholding, editing or calibration decision; the test partition
was read once per cell to produce the reported metrics.

The full protocol---tiers, model roster, tuning budgets, operating rules
and analysis plan---was fixed in version-controlled configuration files
frozen before full-scale execution. No external registry entry was created.

\subsubsection*{Statistical procedures}
The non-inferiority margin was pre-specified at $\delta=0.005$ AUROC before
computation, on three grounds: it lies below any plausibly meaningful
difference at this operating regime, it is roughly one-tenth of the leakage
effect that motivates the study, and it is of the order of the between-seed
standard deviation. Primary comparisons are one-sided non-inferiority tests
and two-one-sided equivalence tests derived from paired DeLong variances on
the seed-0 test partition, under Holm correction. A permutation audit of
the DeLong implementation gave a 6.0\% type-I rate at $\alpha=0.05$ over
400 null draws, within Monte-Carlo error (SE $\approx$1.2\%). Seed-level
comparisons across five paired replications are reported descriptively,
since five replications afford no meaningful hypothesis-testing power.
Uncertainty on between-arm fairness gap differences uses both paired
seed-level contrasts and a 1000-draw stratified (sex$\times$outcome)
bootstrap of the seed-0 test partition.

\subsubsection*{Fairness and explanation details}
Sex is the BRFSS binary SEXVAR. Audits over age bands and race/ethnicity
use the threshold-based arms. Shape-repair edits act on the fitted additive
model: the sex main effect and all sex-involving interaction terms are
zeroed or attenuated ($\gamma=0.5$), and the intercept-equalisation variant
adds a per-group offset $c_g$ selected on validation, giving effective
per-group thresholds $t'_g=\sigma(\sigma^{-1}(t)-c_g)$, which are reported
alongside the arm. Every edit is a pure, reversible function of the fitted
model and is exported as a machine-readable edit log. In the faithfulness
analysis, LIME was additionally applied to the glass-box model itself, with
integer-encoded categoricals, to separate estimator error from cross-model
disagreement.

\subsubsection*{Software and hardware}
All experiments ran on a single workstation with an AMD Ryzen 9 7950X
(16 cores, 32 threads), 128\,GB RAM and one NVIDIA GeForce RTX 4090
(24\,GB VRAM), under PyTorch 2.6.0+cu124 and Python 3.12. Foundation-model
inference used the GPU; all other models used the CPU. Reported
foundation-model inference times use the libraries' default settings, with
no batching tuning, quantisation or context caching, so they represent
out-of-the-box deployment cost rather than an optimised lower bound. The
principal libraries are interpret, XGBoost, LightGBM, CatBoost,
scikit-learn, PyTorch, tabpfn (v2 line), tabicl, Optuna, SHAP, LIME and
imbalanced-learn; versions are pinned in the code release and frozen at run
time. Every result row carries its stage, tier, model, seed and split, and
all tables and figures in this article are generated directly from a single
append-only results file, so no reported number was transcribed by hand.

\subsection*{Supplementary Note}
\suppfig{Supplementary Note — Reporting}
The completed TRIPOD+AI checklist accompanies this submission as a
separate file. The full analysis pipeline, configurations, per-cell
results, and model edit logs are openly available (see Code
availability in the main text).

\end{document}